\documentclass[letterpaper, 10 pt, conference]{ieeeconf}  

\IEEEoverridecommandlockouts                              

\usepackage{amsmath} 
\usepackage{amssymb}  
\usepackage{url}
\usepackage{caption}
\usepackage{subcaption}
\usepackage{siunitx}
\usepackage{cite}
\usepackage{pifont}
\usepackage{booktabs}
\usepackage{tabularx}
\usepackage{multirow}
\usepackage{xspace}

\usepackage{tikz}
\usepackage{pgfplots}
\pgfplotsset{compat=1.18}
\usepackage{adjustbox}
\usepackage{graphicx}

\usepackage[nogroupskip,acronyms,nopostdot,style=super,nonumberlist,toc]{glossaries}
\usepackage{cancel}
\usepackage{diagbox}
\usepackage{lipsum}

\newacronym{ai}{AI}{artificial intelligence}%
\newacronym{drl}{DRL}{deep reinforcement learning}
\newacronym{dl}{DL}{deep learning}
\newacronym{ml}{ML}{machine learning}
\newacronym{rl}{RL}{reinforcement learning}
\newacronym{il}{IL}{imitation learning}
\newacronym{ad}{AD}{autonomous driving}
\newacronym{av}{AV}{autonomous vehicle}
\newacronym{dnn}{DNN}{deep neural network}
\newacronym{ann}{ANN}{artificial neural network}
\newacronym{nn}{NN}{neural network}
\newacronym{dqn}{DQN}{deep Q-network}
\newacronym{rnn}{RNN}{recurrent neural network}
\newacronym{rdqn}{RDQN}{recurrent deep Q-network}
\newacronym{ddqn}{DDQN}{double deep Q-network}
\newacronym{marl}{MARL}{multi-agent reinforcement learning}
\newacronym{dmarl}{DMARL}{deep multi-agent reinforcement learning}
\newacronym{mdp}{MDP}{Markov decision process}
\newacronym{mlp}{MLP}{multilayer perceptron}
\newacronym{lstm}{LSTM}{long short-term memory}
\newacronym{mpc}{MPC}{model predictive control}
\newacronym{its}{ITS}{intelligent transportation systems}
\newacronym{ttc}{TTC}{time-to-collision}
\newacronym{ddpg}{DDPG}{deep deterministic policy gradient}
\newacronym{vae}{VAE}{variational auto-encoder}
\newacronym{mas}{MAS}{multi-agent system}
\newacronym{mal}{MAL}{multi-agent learning}
\newacronym{per}{PER}{prioritized experience replay}
\newacronym{a2c}{A2C}{advantage actor critic}
\newacronym{sg}{SG}{stochastic game}
\newacronym{mg}{MG}{Markov game}
\newacronym{pomdp}{POMDP}{partially observable Markov decision process}
\newacronym{pomg}{POMG}{partially observable Markov game}
\newacronym{dpomdp}{dec-POMDP}{decentralized partially observable Markov decision process}
\newacronym{nrmse}{NRMSE}{normalized root-mean-square error}
\newacronym{ppo}{PPO}{proximal policy optimization}
\newacronym{gae}{GAE}{generalized advantage estimate}
\newacronym{rpl}{RPL}{residual policy learning}
\newacronym{cnn}{CNN}{convolutional neural network}
\newacronym{bev}{BEV}{bird's-eye-view}
\newacronym{fv}{FV}{frontal view}
\newacronym{gt}{GT}{ground truth}

\title{\LARGE \bf
Efficient Learning of Urban Driving Policies Using\\Bird's-Eye-View State Representations
}

\author{Raphael Trumpp$^{1,\star}$, Martin B\"uchner$^{2,\star}$, Abhinav Valada$^{2}$, and Marco Caccamo$^{1}$
\thanks{$\star$ These authors contributed equally.}
\thanks{$^{1}$ TUM School of Engineering and Design, Technical University of Munich, Germany.}%
\thanks{$^{2}$ Department of Computer Science, University of Freiburg, Germany.}
\noindent\thanks{Marco Caccamo was supported by an Alexander von Humboldt Professorship endowed by the German Federal Ministry of Education and Research. Abhinav Valada was supported by the German Research Foundation (DFG) Emmy Noether Program grant number 468878300.}
}%

\begin{document}

\maketitle
\thispagestyle{empty}
\pagestyle{empty}

\begin{abstract}
Autonomous driving involves complex decision-making in highly interactive environments, requiring thoughtful negotiation with other traffic participants. While reinforcement learning provides a way to learn such interaction behavior, efficient learning critically depends on scalable state representations. Contrary to imitation learning methods, high-dimensional state representations still constitute a major bottleneck for deep reinforcement learning methods in autonomous driving. In this paper, we study the challenges of constructing bird's-eye-view representations for autonomous driving and propose a recurrent learning architecture for long-horizon driving. Our PPO-based approach, called RecurrDriveNet, is demonstrated on a simulated autonomous driving task in CARLA, where it outperforms traditional frame-stacking methods while only requiring one million experiences for efficient training. RecurrDriveNet causes less than one infraction per driven kilometer by interacting safely with other road users.
\end{abstract}

\section{Introduction}

After years of research in \gls*{ad} and substantial advances in computer vision, \glspl*{av} are on the verge of becoming a reality. The widespread adoption of \glspl*{av} will change the nature of traffic not only due to increased safety but also in other aspects, such as the inherent decision-making process that now involves human and non-human parties. Driving is a highly interactive task between road users requiring negotiation of all agents. Well-studied in game theory, the agents need to find a cooperative strategy in order to circumvent catastrophic outcomes, e.g., collisions with fatalities. As discussed in \cite{millard2018pedestrians, trumpp2022modeling}, the widespread use of \glspl*{av} will change the perceived risk of human road users since they know that the \glspl*{av} will stop if necessary, therefore leading to inefficient traffic flow. Hence, \glspl*{av} must participate in the negotiation process actively, e.g., letting pedestrians pass a crosswalk but not encouraging pedestrians to jaywalk. 

While \gls*{drl} provides a way to learn such interactive behavior, effective learning critically depends on high-fidelity simulation environments as well as sound state representations used for learning a driving policy. Current \gls*{ad} benchmarks \cite{dosovitskiy17carla} are dominated by \gls*{il} approaches that are easier to design, inspect and maintain since human-level driver data is readily available using auto-pilots \cite{dosovitskiy17carla} or compiled datasets \cite{codevilla2019exploring}. On the contrary, \gls*{drl} suffers from low sample efficiency and requires meaningful, mostly low-dimensional state representations to succeed. We argue that only \gls*{drl} is truly able to reason about interactions in a long-horizon manner and learn from experience, while \gls*{il} will usually not exceed human-level performance. Moreover, \gls*{il} performance may even deteriorate when negotiating with humans in real-world driving scenarios~\cite{millard2018pedestrians}.

\begin{figure}[t]
\centering
\includegraphics[width=8.5cm]{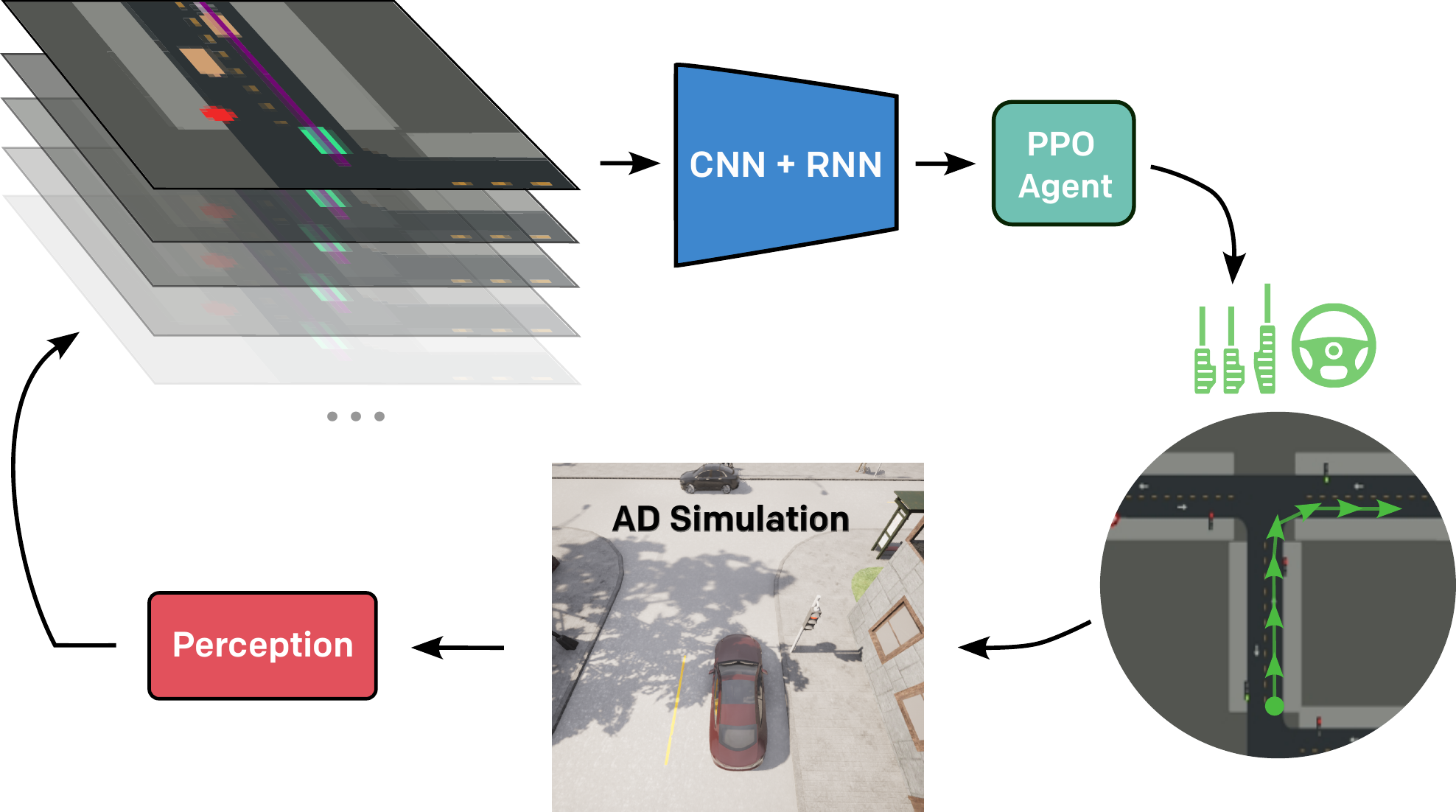}
\caption{Our approach takes BEV images to learn driving policies in a simulated CARLA environment. Utilizing a CNN-based encoder while ensuring temporal understanding via an RNN, we learn longitudinal control using a PPO agent that predicts throttle and braking commands.}
\label{fig:covergirl}
\vspace{-0.7cm}
\end{figure}

In this work, we discuss the challenges of constructing state representations for \gls*{drl} of \gls*{ad} policies and present an efficient recurrent architecture as shown in Fig. \ref{fig:covergirl} that exceeds previous methods based on a number of design choices. As demonstrated in previous approaches by Chen~et~al.~\cite{chen2019free, chen2019deep}, \gls*{drl}-based driving policies are highly dependent on low-dimensional state representations. Throughout this work, we investigate various semantic \gls*{bev} representations used for describing driving states. \gls*{bev} representations may contain vehicles and pedestrians, planned high-level navigation routes, environment maps, and traffic light states. They serve as a highly capable representation while scaling to multiple agents \cite{chen2019free} compared to classical approaches \cite{dosovitskiy17carla}. In addition, we argue that a significant number of perception tasks in \gls*{ad} are tackled in the \gls*{bev} space, such as object detection \cite{huang2021bevdet}, 
segmentation \cite{gosalan21, gosala2023skyeye}, mapping \cite{buchner2023learning} as well as tracking and forecasting \cite{hu2021fiery}. Thus, addressing the \gls*{ad} task in \gls*{bev} benefits from a considerable share of research in the aforementioned domains. In the following, we present an analysis of different \gls*{bev} representations and propose RecurrDriveNet, an efficient recurrent learning architecture that captures the relevant features of the environment while reducing computational complexity. The novelty of RecurrDriveNet is that we use a \gls*{lstm}-based encoding of a series of state transitions, which propagates the hidden state along full trajectories. We demonstrate its driving capability on a simulated \gls{ad} task in CARLA~\cite{dosovitskiy17carla}, where it outperforms traditional learning paradigms that rely on frame-stacking.

Our main contributions can be summarized as follows:
\begin{itemize}
    \item A novel recurrent architecture for efficient \gls{drl} in \gls*{ad} based on semantic \gls*{bev} maps that minimizes collisions with other road users and respects traffic rules.
    \item Extensive experiments and ablations on variations of the \gls*{bev} representation.
    \item We publish our code as part of this work: \small{\url{http://learning2drive.cs.uni-freiburg.de}}.
\end{itemize}
\section{Related Work}

Learning control policies for \gls*{ad} either from demonstration or via reward design proves to be a challenging endeavor. This is not only due to the complexity of actions to take but also errors generated in the perception stage. Existing methods have taken different approaches to modeling state representations while producing low-dimensional intermediate representations. While it is possible to use raw sensor inputs obtained from real-world measurements \cite{kendall2019learning}, the vast majority of relevant works utilize CARLA for synthetic and pseudo-realistic data generation \cite{dosovitskiy17carla, chen2019free, agarwal2021learning}. 

In general, we observe impressive performance in the area of representation learning from raw sensory inputs, namely RGB camera \gls*{fv} and LiDAR \gls*{bev} to be used for decision-making \cite{chen2020reps, prakash2021multi, cheating2020, chen2021interpretable}. The majority of works approach the driving task in an end-to-end manner using RGB camera image \gls*{fv} and/or LiDAR \gls*{bev}. Interestingly, the larger the perception backbone, the more we see approaches opting for \gls*{il} \cite{chen2022allveh, prakash2021multi, chen2019deep, shao2023interfuser} instead of \gls*{drl}. Thus, efficient state representations are still a considerable bottleneck when applying \gls*{drl} to the problem at hand. Based on this, we classify related works based on the chosen state representations and modalities in the following.

The original CARLA paper introduced a robust, modular method with distinct modules for perception, planning, and control. It uses stacked $84\times 84$ \gls*{fv} images, a state vector including current speed, distance to the goal, and high-level navigation commands \cite{dosovitskiy17carla}.
Some works regress driving-related affordances from temporal stacks of RGB \gls*{fv} images \cite{survey_e2eAD} that are fed to low-level controllers to infer commands, which effectively induces human bias in the system. 

With the goal of incorporating multiple modalities, Chen~et~al. take a representation learning approach to generate pseudo-synthetic semantic \gls*{bev} representations in an auto-encoding scheme from an RGB \gls*{fv} image as well as LiDAR data~\cite{chen2021interpretable, chen2020reps, chen2019free}. The produced \gls*{bev} representations combine several map layers, namely drivable surface, planned route, stacked actor history, and stacked ego history, all at $64\times 64$ resolution. Even though these are prone to errors, they provide high interpretability, which is vital to the \gls*{ad} task. Interestingly, Chen et al. \cite{cheating2020} take a knowledge distillation approach by first training a privileged teacher agent that has access to the \gls*{gt}, which is in turn used to train a real-world student agent that takes only RGB \gls*{fv} images as input and is supervised by the teacher. Inspired by these advances, a number of approaches directly use \gls*{gt} information to synthetically generate \gls*{bev} representations instead of obtaining these from raw sensor inputs. An \gls*{il} approach by Chen et al. \cite{chen2019deep} uses \gls*{bev} images of size $192\times 192$ that also include traffic light information along the planned route and stacked actor history. A \gls*{drl} pendant to the aforementioned method \cite{chen2019free} uses a smaller resolution of $64\times 64$ as high-dimensional state spaces still pose a considerable hurdle for \gls*{drl}. Similarly, others \cite{chen2019free,agarwal2021learning} opt for temporal stacking of \gls*{bev} maps of resolution $128\times 128$~\cite{agarwal2021learning}. Notably, all these approaches do not make use of recurrent approaches in favor of frame-stacking.

Contrary to the \gls*{bev} representation, Prakash et al. \cite{prakash2021multi} learn multi-modal fusion with transformers from \gls*{fv} and LiDAR \gls*{bev}. Similarly, Shao et al. \cite{shao2023interfuser} also include RGB surround view images to predict future waypoints auto-regressively.
In contrast to this, \gls*{drl} approaches usually predict driving control commands such as steering angle, throttle, and braking. 

Most similar to our work, both Chen et al. \cite{chen2019free} and Agarwal et al. \cite{agarwal2021learning} use semantic \gls*{bev} state representations with access to \gls*{gt} information. While we benchmark different \gls*{bev} representations for learning efficiency in the context of \gls*{drl}, the focus of our work is the introduction of a recurrent learning scheme that propagates the hidden state along all steps of the observed trajectory. Alleviating the need for frame-stacking architectures ultimately leads to improved agent behavior given the chosen reward objective.

\section{Background}
\label{seq:background}

\subsection{Deep Reinforcement Learning for Autonomous Driving}
\label{subseq:drl_ad}
The goal of \gls*{drl} is to maximize a reward signal by learning the weights $\theta$ of an agent's (stochastic) policy $\pi_{\theta}(a_t | s_t)$ that maps actions $a_t$ to an observed state $s_t \in \mathcal{S}$. In \gls*{ad}, the state of the ego vehicle is typically defined by its own position,  velocity, and planned trajectory as well as the state of other agents participating in the driving scenario. When action $a_t$ is applied, the agent observes a new state $s_{t+1}$ according to the state transitions probability $\mathcal{T}$. The function $\mathcal{R}$ defines the reward signal $r_t$; the discount factor $\gamma$ weights present to future rewards. Given an environment where an ego vehicle learns to interact with $N$ other road users, the system can then be described by a \gls{mdp} with the tuple $(\mathcal{S}, \mathcal{A}, \mathcal{T}, \mathcal{R}, \gamma)$ modeling the other road users as part of the environment.

Since interaction-rich scenarios in \gls*{ad} involve $N\neq\operatorname{const}$ participating road users, a count-invariant state representation $s_t \in \mathcal{S}$ has to be found. This is due to the fact that the \gls*{dnn} used in \gls*{drl} require an input tensor of fixed size. Therefore, a common approach is to use a \gls*{bev}-based state representation that encodes the state as an image. However, such image representations encode only spatial information. Due to the dynamic nature of interactions in \gls*{ad}, \gls{drl} agents must observe first and second-order derivatives of the vehicle dynamics too, i.e., velocity and acceleration. This information must either be encoded into ${s}_t$ directly or by opting for a temporal representation.

\subsection{Deep Reinforcement Learning Algorithm}
\label{subseq:drl}
\Gls*{ppo}~\cite{schulman2017proximal} is an \textit{on}-policy \gls*{drl} algorithm using \glspl*{dnn} to learn a stochastic policy $\pi_{\theta_k}(a_t | s_t)$. The \gls*{dnn}'s weights $\theta_k$ are updated by
\begin{equation}
\label{eqn:ppo_update}
\theta_{k+1}=\arg \max _\theta \underset{s_t, a_t \sim \pi_{\theta_k}}{\mathrm{E}}\left[L\left(s_t, a_t, \theta_k, \theta\right)\right],
\end{equation}
at iteration $k+1$ for transition tuples $e_t=\{s_t, a_t, r_{t},s_{t+1}\}$ of a $T$ fixed-sized steps long trajectory $\tau=\{e_0, e_1, ..., e_T\}$. The objective function $L\left(s_t, a_t, \theta_k, \theta\right)$ is defined as 
\begin{align}
    L\left(s_t, a_t, \theta_k, \theta\right) = \min \biggl( &\frac{\pi_\theta(a_t \mid s_t)}{\pi_{\theta_k}(a_t \mid s_t)} A^{\pi_{\theta_k}}(s_t, a_t), \nonumber \\ 
     & g\left(\epsilon, A^{\pi_{\theta_k}}(s_t, a_t)\right) \biggr).
\end{align}
The advantage function $A=A^{\pi_{\theta_k}}(s_t, a_t)$ is used to calculate 
\begin{equation}
\label{eqn:g_eps}
g(\epsilon, A)= \begin{cases}(1+\epsilon) A & A \geq 0 \\ (1-\epsilon) A & A<0 \end{cases},
\end{equation}
with the hyperparameter $\epsilon$. Due to $\epsilon$ and the min-clipping in (\ref{eqn:ppo_update}), the policy updates are regularized to ensure smooth policy updates. To calculate the advantage function $A$ for (\ref{eqn:g_eps}), a common choice is the \gls*{gae}~\cite{schulman2015high}. This method is based on a value function $V_{\phi_k}(s_t)$ with weights $\phi_k$ that is learned based on
\begin{equation}
    \phi_{k+1}=\arg \min _\phi \frac{1}{\left|\mathcal{D}_k\right| T} \sum_{\tau \in D_k} \sum_{t=0}^T\left(V_{\phi_k}\left(s_t\right)-\hat{R}_t\right)^2,
\end{equation}
with the reward-to-go $\hat{R}_t=\sum_{t'=t}^{T}r_t$. The final regression target is obtained by estimating the expectation in (\ref{eqn:ppo_update}) with $B$ recorded $T$-step long trajectories $\mathcal{D}_k=\{\tau_0, \tau_2, \dots, \tau_{B-1}\}_{k}$.
\section{Methodology}
\label{seq:methodology}

In this work, we present a recurrent agent architecture for efficient \gls*{drl} based on semantic \acrfull{bev} maps, which serve as scalable state representations. We assume access to \gls*{gt} information in order to retrieve actor and environment data, which is in line with related works \cite{agarwal2021learning, chen2019deep}. The use of \gls*{bev} representations is motivated by recent advancements \cite{gosala2023skyeye, buchner2023learning} and the fact that \gls*{bev} images are sensor-agnostic representations. In the following, we describe spatial state representations (Sec.~\ref{subseq:var_len_state_rep}) as well as temporal state representations (Sec.~\ref{subseq:seq_state_rep}) that allow making the agents' dynamics observable as discussed in Sec.~\ref{subseq:drl_ad}.

\begin{figure}[htbp]
\begin{subfigure}{0.22\textwidth}
\centering
\vspace{0.2cm}
\includegraphics[width=0.7\linewidth]{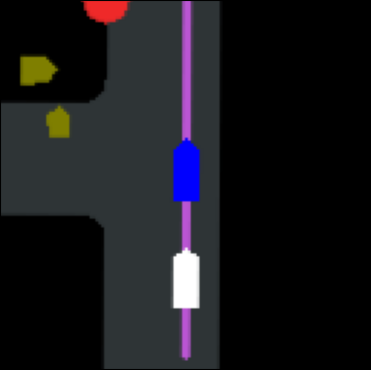} 
\caption{RGB-BEV: Road geometry (gray), future waypoints (pink), traffic light states (green, yellow, red), ego vehicle (white), enlarged pedestrians (dark yellow), other vehicles (blue).}
\label{fig:rgb_bev}
\end{subfigure}
\begin{subfigure}{0.22\textwidth}
\centering
\vspace{0.2cm}
\includegraphics[width=0.7\linewidth]{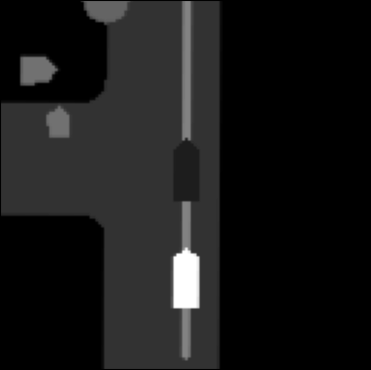}
\caption{Gray-BEV: Objects are encoded using distinct intensities in the gray spectrum that resemble the $\alpha$-channel of (a).\\}
\label{fig:gray_bev}
\end{subfigure}
\vspace{-0.2cm}
\begin{subfigure}{0.48\textwidth}
\centering
\includegraphics[width=0.97\linewidth]{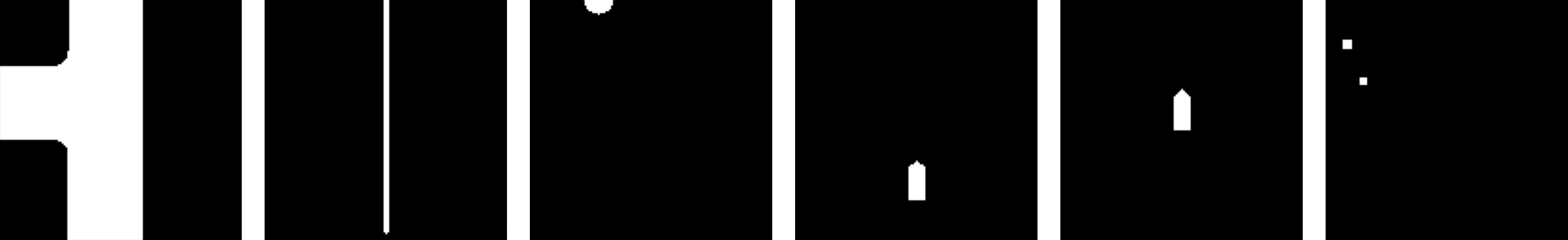}
\caption{Multi-BEV: The channels encode map geometry (\#~1), future waypoints (\#~2), traffic lights (\#~3), ego vehicle (\#~4), other vehicles (\#~5), and pedestrians (\#~6).}
\label{fig:multi_bev}
\end{subfigure}
\vspace{-0.7cm}
\caption{State representations as \glspl*{bev}.}
\label{fig:state_bev}
\vspace{-0.3cm}
\end{figure}

\subsection{Spatial State Representations}
\label{subseq:var_len_state_rep}
We utilize a spatial state $s_t$ that encodes all actors as well as the immediate environment in a \gls*{bev} image of the current driving scene. The pose of all road users is represented in an oriented bounding box fashion as shown in Fig.~\ref{fig:state_bev}.
Environment information is encoded by underlaying a map representation with lane information. As discussed in Sec.~\ref{subseq:drl_ad}, the chosen state representation is invariant to the number of participating road users. In the following, we discuss different variants of the \gls*{bev} representation.

\subsubsection{RGB Bird's Eye View (RGB-BEV)}
The \gls*{bev} is represented by a quadratic image with 3 color channels (RGB) and dimension $3 \times K \times K$. Different object types are color-coded in the image representation as shown in Fig.~\ref{fig:rgb_bev}.

\subsubsection{Grayscale Bird's Eye View (Gray-BEV)}
The RGB-BEV representation is converted to an image with a single channel in grayscale (Fig.~\ref{fig:gray_bev}) but with the same properties, i.e., the representation has dimension $1 \times K \times K$.

\subsubsection{Multi-Channel Bird's Eye View (Multi-BEV)}
Instead of encoding the road user's location into an RGB image, an image representation of dimension $6 \times K \times K$ is used (Fig.~\ref{fig:multi_bev}), with each of the six channels corresponding to a different property, respectively. 

\subsection{Temporal State Representations}
\label{subseq:seq_state_rep}

\begin{figure*}[h]
\begin{subfigure}{0.5\textwidth}
\centering
\vspace{0.2cm}
\includegraphics[width=0.76\linewidth]{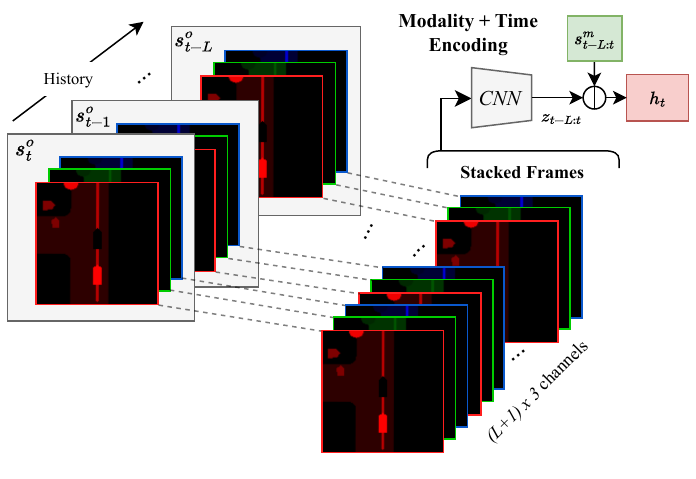} 
\caption{Frame-stacking of sequences as used in \cite{agarwal2021learning, chen2019free}.}
\label{fig:frame_stacking}
\end{subfigure}
\begin{subfigure}{0.5\textwidth}
\centering
\vspace{0.2cm}
\includegraphics[width=0.7\linewidth]{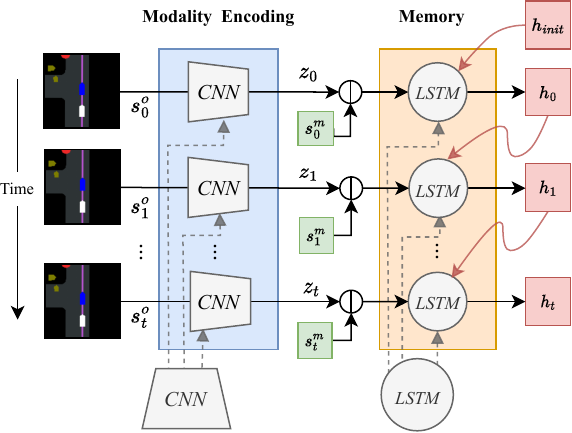}
\caption{LSTM-based encoding of trajectories (ours).}
\label{fig:lstm}
\end{subfigure}
\vspace{-0.5cm}
\caption{Sequential state representations and their encoding: Instead of stacking frames $s^{o}_{t}$ along $L+1$ frames as shown under (a), we encode \gls*{bev} images using a recurrent architecture (b) in order to generate successive hidden states $h_{t}$ used for policy learning.}
\label{fig:image2}
\vspace{-0.3cm}
\end{figure*}

The \gls{bev} state representation in Sec.~\ref{subseq:var_len_state_rep} encodes the road users' location into an image. However, the pure observation of such an image does not yield further insights into the road users\ states regarding linear velocity and acceleration.
To make these properties observable, the frames can either be trivially stacked to form a history of previous observations or by employing recurrency along the trajectory of state transitions as we do (see Sec.~\ref{subsubsec:rnn_rep}).

\subsubsection{Frame-Stacking}
\label{subsubsec:frame_stacking}
The easiest way to include past information into the current state $s_t$ is to stack past observed states ${s_{t-1}, s_{t-2}, ..., s_{t-L}}$ in terms of \gls*{bev} images, i.e., the RGB-BEV representation becomes an $((L+1) \cdot 3) \times K \times K$ image, forming a history of the $L$ previous states as shown in Fig.~\ref{fig:frame_stacking}. This history allows the \gls*{drl} agents to observe the difference between subsequent states, which makes the other road users' velocity and acceleration observable. This constitutes the method previous works followed \cite{agarwal2021learning, chen2019free}.

\subsubsection{Recurrent Neural Network}
\label{subsubsec:rnn_rep}
Instead of encoding the history into the state representation itself, the use of \glspl*{rnn} allows the \gls*{drl} agent to build a hidden state $h_t$ that is updated at every observed state $s_t$. The updated hidden state is accessed by the agent at the next time step again. This allows the agent to keep a memory of past observed states. In related works, \glspl*{rnn} are typically used to only encode the history of a sequence of frame-stacked images as described in Sec.~\ref{subsubsec:frame_stacking}. Our presented architecture makes use of the full long-horizon capabilities of \glspl*{lstm} by propagating the hidden state along the full trajectory of states as shown in Fig.~\ref{fig:lstm}.

\subsection{RecurrDriveNet and Baselines}
Based on the state representations outlined in Sec.~\ref{subseq:var_len_state_rep} and \ref{subseq:seq_state_rep}, we present the model of an \gls*{av} that learns to drive in a simulated environment in CARLA \cite{dosovitskiy17carla}. This ego vehicle is implemented as a \gls*{ppo} agent consisting of a policy network $\pi_{\theta}$ and a value network $V_{\phi}$. In this context, we introduce our agent model called RecurrDriveNet, which combines the Multi-\gls*{bev} state representation (Sec.~\ref{subseq:var_len_state_rep}) with an \gls*{lstm} architecture. The novelty of RecurrDriveNet is that the hidden state $h_t$ is constantly propagated along each state of a trajectory (Fig.~\ref{fig:lstm}) allowing for efficient long-horizon reasoning. This approach is only feasible because \gls*{ppo} is an \textit{on}-policy algorithm that requires rollouts of trajectories. In Sec.~\ref{seq:results}, we evaluate our proposed architecture across three different state representations. In the following, we point out the differences between these models.
 
\subsubsection{Action Space}
Based on available map data, the agent has to follow a planned trajectory of waypoints. These waypoints are generated randomly by a global planning module and the map's lane graph. We employ a Stanley controller \cite{hoffmann2007autonomous} for lateral control of the vehicle to simplify the action space of the agent so that only longitudinal control has to be learned. We justify this by arguing that the vehicle-pedestrian interaction in urban areas is mostly dependent on the longitudinal acceleration of the ego vehicle.

The ego vehicle's action space is defined as $a_t \in [-1, 1]$, which maps to the low-level control of a CARLA vehicle in the form of braking and throttle action, i.e., $-1$ means full brake, $0$ idle, and $+1$ corresponds to full throttle.

\subsubsection{Observation Space}
Instead of learning from raw sensor data such as RGB \gls*{fv} camera or LiDAR, we use \gls*{gt} perception for representing the ego-vehicle surroundings including other actors similar to others \cite{agarwal2021learning, chen2019free}. Their position and orientation are then reflected in one of the \gls*{bev} representations discussed in Sec.~\ref{subseq:var_len_state_rep}. 

After encoding the state representation $s_{t}$ using a modality-encoder network (Fig.~\ref{fig:lstm}), we concatenate the encodings $z_t$ with an ego-state vector $s_t^{m}$ that contains the following variables:
\begin{equation}
\label{eqn:StateCar}
s_t^{m} = 
        \Big[
           {\Delta}^{\text{traj}}_{t}, {\delta}^{\text{traj}}_{t}, {{v}}^{\text{ego}}_{t}, {{\dot v}}^{\text{ego}}_{t}, \alpha^{\text{ego}}_{t}, {{v}}^{\text{lim}}_{t}, {{f}}^{\text{red}}_{t}
        \Big]^{\top}.
\end{equation}
The deviation from the planned trajectory is defined by the distance $ {\Delta}^{\text{traj}}_{t}$ to the closest waypoint and the relative orientation ${\delta}^{\text{traj}}_{t}$ to it. The ego vehicle's velocity, acceleration, and steering angle are ${{v}}^{\text{ego}}_{t}$, ${{\dot v}}^{\text{ego}}_{t}$, and $\alpha^{\text{ego}}_{t}$, respectively. The current allowed driving speed is given by ${{v}}^{\text{lim}}_{t}$. When the ego vehicle is close to a red traffic light, ${{f}}^{\text{red}}_{t}$ is set to $1.0$, $0.5$ when yellow, and otherwise to $0$.

\subsubsection{Reward Function}
Modified from \cite{chen2021interpretable}, we define the reward function as:
\begin{align}
    r = & v^{\text{ego}}_{\text{long}} - 10 \cdot f_{s} - 0.2 \cdot |\alpha| (v^{\text{ego}}_{\text{long}})^{2} - 5 \cdot \alpha^2 \nonumber \\
     &  - f_{o} - 200\cdot f_{v} - 200\cdot f_{p} - 200\cdot f_{r},
\end{align}
where each $f_{*}$ is set to 0 when the following corresponding conditions are not true. When the vehicle does not obey the permitted driving speed, $f_{s}$ evaluates to $f_{s}=1$. Excessive use of the steering angle $\alpha$ is penalized together with when the vehicle diverges more than 1$\si[per-mode=symbol]{\m}$ from the planned trajectory by setting $f_{o}=1$. Additionally, collisions with pedestrians and other vehicles lead to a penalty with $f_{v}=1$ or $f_{p}=1$, respectively. When the ego vehicle runs a red traffic light,  we set $f_{r}=1$.

\subsubsection{Network Architecture}
The agent's policy network $\pi_{\theta}$ and the value network $V_{\phi}$ share a common modality-encoder to encode the \gls*{bev} state $s^{o}_{t}$ to $z_t$. We deploy a \gls{cnn} with three convolutional layers $\{(32, 8, 4), (64, 4, 2), (64, 3, 2)\}$ with (\#-filters, size, stride) and ReLU activation including a subsequent linear projection layer to $z_t \in \mathbb{R}^{256}_{+}$. The encodings $z_t$ are concatenated with the measurement vector $s^{m}_{t}$. For the \gls*{lstm}-based agents, these combined encodings are fed to a single \gls*{lstm}-layer with a hidden state of $h_t \in \mathbb{R}^{256}$. The \gls*{lstm}'s output is then used in two separate modules of single linear layers for the policy and value network, respectively. Since the learned policy in \gls*{ppo} is stochastic, the agent learns the parameters of a $\operatorname{tanh}$-Normal distribution, i.e., a Gaussian distribution that is projected using a $\operatorname{tanh}$-function to the interval of $[-1,1]$. When no \gls*{lstm} is used, the combined encoding is directly fed to the two modules with two instead of one linear layer.

\section{Experimental Evaluation}
\label{seq:results}

\subsection{Simulation Setup}
Our proposed agent models are evaluated using the CARLA \gls*{ad} simulator ~\cite{dosovitskiy17carla} that allows for imitating realistic traffic negotiation among both pedestrians and vehicles. We use the \textit{Town02} map, which resembles an urban area with single-lane roads and traffic lights but no roundabouts. Our simulation includes 30 other vehicles and 50 simulated pedestrians across the map that are controlled by CARLA. A LiDAR sensor with a range of $30\,\si[per-mode=symbol]{\m}$, $110\,^\circ$ field of view, and no noise is used; traffic participants can be occluded. The action frequency is at $10\,\si[per-mode=symbol]{\Hz}$.

\subsection{Implementation and Training Details}
We train a total of six architectures for \gls{drl} in \gls{ad}, as detailed in Table~\ref{tab:state_representations} and Table~\ref{tab:results-successor}. 
Our proposed RecurrDriveNet is based on the Multi-\gls*{bev} state $s_t^{\text{Multi}}$ combined with an \gls*{lstm} for recurrent connection.
For all agents, the \gls*{bev} image is a square image with $K=128$ pixels. The modality encoder's first convolutional layer with $\text{stride}=4$ allows the use of high-resolution input images without increasing the output dimension significantly. Lower-resolution images suffer from ambiguity of the bounding box orientation, especially for pedestrians. The frame-stacking variants take the current and the previous $L=4$ observed frames as input.

All models are optimized for $1\,e\textsuperscript{6}$ steps with the same set of hyperparameters; reward and observations are normalized by a running mean calculation. We select $\gamma=0.999$, a clipping value of $\epsilon=0.1$ for \gls{ppo}, and collect trajectories with a length of $128$ steps with four CARLA environments running in parallel. Further implementation details are accessible in our published code repository.
\begin{table}
\centering
\scriptsize
\caption{Combinations of state representations evaluated in this work. RecurrDriveNet is based on the state representation ${{s}_{t}^{\text{Multi}}+h_{t-1}}$.}
\begin{tabular}{c|cr|cr}
\toprule
\multirow{2}{*}{\diagbox[width=7em]{Modality}{Sequence}} & \multicolumn{2}{c|}{LSTM} & \multicolumn{2}{c}{{Frame Stacking}} \\ & State $s_{t}$ & dim & State  $s_{t}$ & dim \\
\midrule
{RGB-BEV} & ${s}_{t}^{\text{RGB}}+h_{t-1}$ & $3 \times 128^2$ & ${s}_{t-L:t}^{\text{RGB}}$ & $15 \times 128^2$ \\
{Gray-BEV} & ${s}_{t}^{\text{Gray}}+h_{t-1}$ & $1 \times 128^2$ & ${s}_{t-L:t}^{\text{Gray}}$ & $5 \times 128^2$ \\
Multi-BEV & ${s}_{t}^{\text{Multi}}+h_{t-1}$ & $6 \times 128^2$ & ${s}_{t-L:t}^{\text{Multi}}$ & $30 \times 128^2$\\
\bottomrule
\end{tabular}
\label{tab:state_representations}
\vspace{-0.2cm}
\end{table}

\subsection{Evaluation Metrics}
The performance of the presented models is evaluated based on the occurrence of infractions while driving:
\begin{enumerate}
    \item $I_{\text{veh}}$: \# of vehicle collisions per distance in \si[per-mode=symbol]{1\per\km}.
    \item $I_{\text{ped}}$: \# of pedestrians collisions over distance in \si[per-mode=symbol]{1\per\km}.
    \item $I_{\text{red}}$: \# of run red traffic lights over distance in \si[per-mode=symbol]{1\per\km}.
\end{enumerate}
We further combine these in a metric termed $I_{\Sigma}$, which represents the sum across all infraction scores. The average violation with respect to the speed limit $v_{\text{lim}}^{\text{dev}}$ is given in \si[per-mode=symbol]{\%}. Additionally, we calculate the velocity $v^{\text{ego}}_{\text{move}}$ in \si[per-mode=symbol]{\m\per\s} when the ego-vehicle is moving faster than $0.2\,\si[per-mode=symbol]{\m\per\s}$.

The results are evaluated on 320k simulation steps corresponding to a real-world driving time of almost 9\,h.

\subsection{Training Results and Qualitative Assessment}
\begin{figure}[t]
    \centering
    \vspace{0.2cm}
    \includegraphics[width=0.44\textwidth]{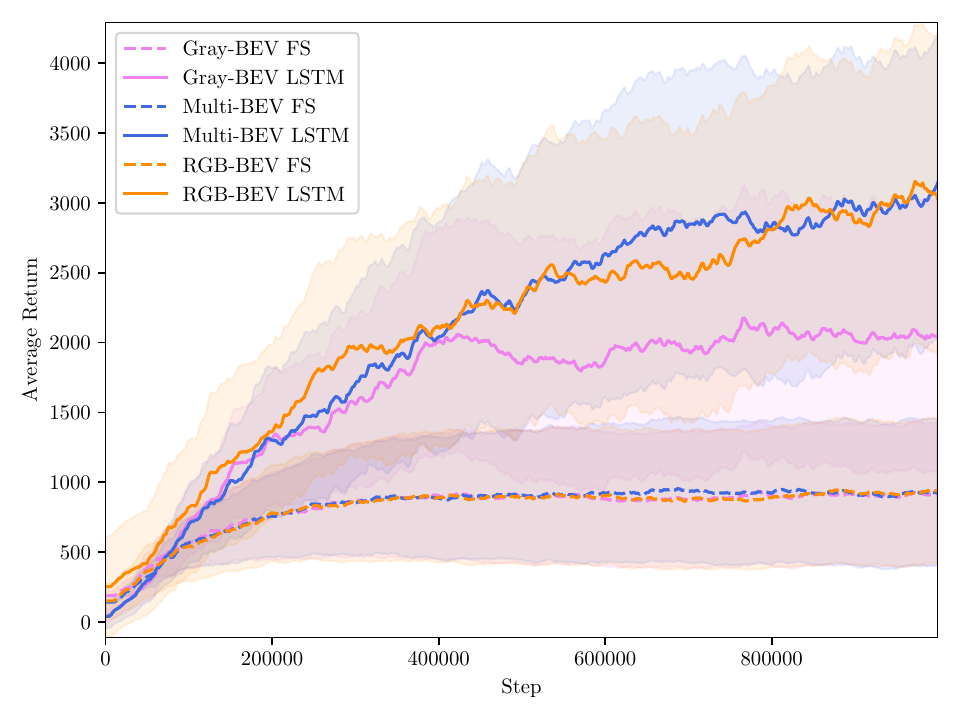}
    \caption{Comparison of the average returns produced by all model variants over the training time. We average across 3 runs per variant and display the associated standard deviation.}
    \label{fig:avg_return}
    \vspace{-0.5cm}
\end{figure}

As shown in Figure~\ref{fig:avg_return}, only the \gls*{lstm}-based agents converge to a high average episodic return above 3000. We hypothesize that the models that use frame-stacking are stuck in local minima; they lack the long-horizon reasoning capability of the \gls*{lstm} models. The Gray-\gls{bev} \gls{lstm} variant is superior to the frame-stacking approaches but does not achieve the high performance of the Multi-\gls{bev} LSTM (RecurrDriveNet) and RGB-\gls{bev} LSTM models.
Despite similar training performance, further qualitative analysis shows that RecurrDriveNet's Multi-\gls{bev} LSTM model outperforms the RGB-\gls{bev} LSTM variant by preventing infractions more reliably.
We observe that all agents stop at large distances from other vehicles. While this behavior may be effective in preventing collisions, it leads to non-optimal traffic flow.

\begin{table*}
\centering
\scriptsize
\setlength\tabcolsep{3.7pt}
\caption{Quantitative results of our LSTM-based RecurrDriveNet for different state representations (Sec.~\ref{seq:methodology}) and comparison against traditional frame-stacking representations used by Chen et al. \cite{chen2019free} and Agarwal et al. \cite{agarwal2021learning}. The displayed results represent each variant's best results across 3 training runs.}
\begin{tabular}{p{2.3cm} p{1.2cm} p{0.5cm} p{0.5cm} | p{1.4cm} | p{1.4cm} p{1.4cm}  p{1.4cm} | p{1.3cm} p{1.3cm} p{1.3cm}}
 \toprule
 Model &  State Repr. & \centering{LSTM} & \centering{FS} & $I_{\Sigma} \downarrow$ \scriptsize{[\si[per-mode=fraction]{\per\km}]} & $I_{\text{veh}}\downarrow$ \scriptsize{[\si[per-mode=fraction]{\per\km}]}  & $I_{\text{ped}}\downarrow$ \scriptsize{[\si[per-mode=fraction]{\per\km}]} & $I_{\text{red}} \downarrow$ \scriptsize{[\si[per-mode=fraction]{\per\km}]}  & $v_{\text{lim}}^{\text{dev}}\downarrow$ \scriptsize{[\si[per-mode=fraction]{\%}]} & $v^{\text{ego}}_{\text{move}}$ \scriptsize{[\si[per-mode=fraction]{\meter\per\second}]}  \\
 \midrule
 \tiny
    & Gray-BEV &   & \centering{\checkmark} & $8.68$  & $3.35$ & $\boldsymbol{0.44}$ & $4.89$ & $2.93$ &  $7.00$   \\
    
    & RGB-BEV &  & \centering{\checkmark} & $8.30$  & $2.86$ & $0.46$ & $4.98$ & $3.02$ & $6.95$  \\

    & Multi-BEV & & \centering{\checkmark} & $8.47$ &  $2.92$ & $0.51$ & $5.03$  & $1.26$ & $6.82$  \\
    
    & Gray-BEV & \centering{\checkmark} & & $3.65$ & $2.90$ & $0.67$ & $0.08$  & $2.78$ & $5.85$  \\
    
    & RGB-BEV & \centering{\checkmark} & & $1.24$ & $0.25$ & $0.93$ & $\boldsymbol{0.06}$ & $0.76$ & $4.66$  \\
    
 \midrule
 RecurrDriveNet (ours)  &  Multi-BEV & \centering{\checkmark}  & & $\boldsymbol{0.75}$  & $\boldsymbol{0.15}$ & $0.54$ & $0.07$  & $1.41$ & $4.67$  \\
 \bottomrule
\end{tabular}
\label{tab:results-successor}
 \end{table*}

\subsection{Infraction Analysis}

Table \ref{tab:results-successor} presents the results of RecurrDriveNet compared to classical frame-stacking representations as used in previous works \cite{agarwal2021learning, chen2019free}.
First, we observe that only the \gls*{lstm}-based architectures are able to learn a driving policy that is able to stop at red traffic lights. Second, RecurrDriveNet estimates the dynamic behavior of other road users successfully in order to prevent infractions as it achieves the lowest overall infraction score $I_{\Sigma}$ with a value of $0.75$ per driven kilometer. While all \gls*{lstm}-based agents excel at obeying red traffic lights, i.e., infraction values $I_{\text{red}}$ of $0.06$, $0.07$, and $0.08$ for the RGB-\gls{bev}, Multi-\gls{bev}, and Gray-\gls{bev} state representations, respectively, only the Multi-\gls{bev}-based RecurrDriveNet has learned to reduce vehicle collisions to a minimum of $I_{\text{veh}}=0.15$. We reason that the Multi-\gls{bev} state has the advantage that each channel's property is clearly separated from the remaining ones, i.e., a vehicle does not have to be detected across multiple channels. Similarly, the Gray-\gls{bev} is of much lower dimension compared to RGB-\gls{bev}. Interestingly, the pedestrian infractions of our RecurrDriveNet remain at a similar level as the other models with $I_{\text{veh}}=0.54$. This indicates that the \gls{bev} image representation struggles with representing pedestrian dynamics, e.g., probably due to smaller bounding boxes in the image and abrupt changes in the walking direction. The low performance of the frame-stacking representations corresponds to findings in previous work where frame-stacking approaches tend to work but require long training time, e.g., Agarwal et al. \cite{agarwal2021learning} trained their network for a total of $12\,e\textsuperscript{6}$ time steps compared to our $1e\,\textsuperscript{6}$ steps. Although RecurrDriveNet exhibits safe driving behavior, its driving is not overly cautious as a high moving velocity $v^{\text{ego}}_{\text{move}}=4.67\,\si[per-mode=fraction]{\meter\per\second}$ shows. Eventually, all agents comply with the set speed limits given by $v_{\text{lim}}^{\text{dev}}$.

\section{Conclusion}
\label{seq:conclusion}

While providing an extensive efficiency comparison of \gls*{bev} representations used for \gls*{ad}, we presented a novel recurrent approach to learning urban driving policies from \glspl*{bev}. The introduced RecurrDriveNet agent reduces collisions with other vehicles to a minimum while obeying traffic rules. Crucial to this approach is a \gls*{lstm}-based learning paradigm for solving long-horizon driving. Our experiments show that this method surpasses frame-stacking-based methods used in the past. Follow-up work could address the incorporation of trajectory prediction methods in the \gls*{bev} representation. The potential effects of the reward design on more efficient traffic flow should also be examined.

\bibliographystyle{IEEEtran}
\bibliography{references.bib}

\end{document}